\documentclass[letter,12pt]{article}
\usepackage{arxiv}

\usepackage[utf8]{inputenc} 
\usepackage[T1]{fontenc}    
\usepackage{hyperref}       
\usepackage{url}            
\usepackage{booktabs}       
\usepackage{amsfonts}       
\usepackage{nicefrac}       
\usepackage{microtype}      
\usepackage{lipsum}		
\usepackage{graphicx}
\usepackage{natbib}
\usepackage{doi}
\usepackage{caption}
\usepackage{multirow}
\usepackage{makecell}
\usepackage{subcaption}
\usepackage{amsmath}

\usepackage{mathptmx}
\usepackage[T1]{fontenc}

\title{Automatic Text Simplification of News Articles in the Context of Public Broadcasting}

\author{{Diego Maupomé, Fanny Rancourt, Thomas Soulas, Alexandre Lachance, \href{https://orcid.org/0000-0001-8196-2153}{\includegraphics[scale=0.06]{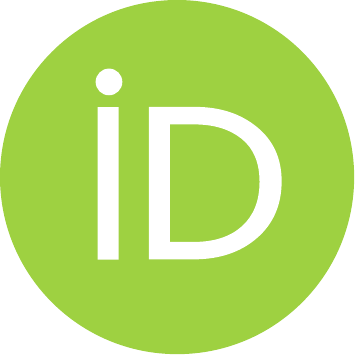}\hspace{1mm}Marie-Jean Meurs}}\\
	Université du Québec à Montréal\\
	Montreal, QC, Canada\\
	\texttt{\{maupome.diego,rancourt.fanny.2,soulas.thomas\_david,lachance.alexandre.5\}@courrier.uqam.ca}\\
    \texttt{meurs.marie-jean@uqam.ca} \\
    \AND {Desislava Aleksandrova, Olivier Brochu Dufour, Igor Pontes}\\
	   CBC/Radio-Canada\\
	   Montréal, QC, Canada\\
    \texttt{\{dessy.aleksandrova, olivier.brochu.dufour\}@radio-canada.ca} \\
	\AND
	\href{https://orcid.org/0000-0002-7858-3504}{\includegraphics[scale=0.06]{orcid.pdf}\hspace{1mm}Rémi Cardon} \\
	Université catholique de Louvain\\
	Louvain-la-Neuve, Belgique\\
	\texttt{remi.cardon@uclouvain.be} \\
    \And{Michel Simard, Sowmya Vajjala}\\
    Conseil National de Recherche du Canada\\
    Ottawa, Canada\\
	\texttt{\{michel.simard,sowmya.vajjala\}@nrc-cnrc.gc.ca}\\
}

\date{}

\renewcommand{\shorttitle}{Automatic Text Simplification of News Articles}

\hypersetup{
pdftitle={Automatic Text Simplification of News Articles in the Context of Public Broadcasting},
pdfsubject={cs.CL},
pdfauthor={Diego Maupomé, Fanny Rancourt, Thomas Soulas, Alexandre Lachance, Marie-Jean Meurs, Desislava Aleksandrova, Olivier Brochu Dufour, Igor Pontes, Rémi Cardon, Michel Simard, Sowmya Vajjala},
pdfkeywords={Automatic Text Simplification, News Articles, Public Broadcasting, Natural Language Processing},
}

\begin{document}
\maketitle



\section{Introduction}

This report summarizes the work carried out by the authors during the \textit{Twelfth Montreal Industrial Problem Solving Workshop}, held at Université de Montréal in August 2022. 
The team tackled a problem submitted by CBC/Radio-Canada on the theme of \textit{Automatic Text Simplification} (ATS). 
In order to make its written content more widely accessible, and to support its second-language teaching activities, CBC/RC has recently been exploring the potential of automatic methods to simplify texts. 
They have developed a modular \textit{lexical simplification system} (LSS), which identifies complex words in French and English texts, and replaces them with simpler, more common equivalents. 
Recently however, the ATS research community has proposed a number of approaches that rely on deep learning methods to perform more elaborate transformations, not limited to just lexical substitutions, but covering syntactic restructuring and conceptual simplifications as well. 
The main goal of CBC/RC’s participation in the workshop was to examine these new methods and to compare their performance to that of their own LSS.

This report is structured as follows: In Section~\ref{sec:pb_desc}, we detail the context of the proposed problem and the requirements of the sponsor. We then give an overview of current ATS methods in Section~\ref{sec:rel_work}. Section~\ref{sec:datasets} provides information about the relevant datasets available, both for training and testing ATS methods. As is often the case in natural language processing applications, there is much less data available to support ATS in French than in English; therefore, we also discuss in that section the possibility of automatically translating English resources into French, as a means of supplementing the French data. The outcome of text simplification, whether automatic or not, is notoriously difficult to evaluate objectively; in Section~\ref{sec:eval}, we discuss the various evaluation methods we have considered, both manual and automatic. Finally, we present the ATS methods we have tested and the outcome of their evaluation in Section~\ref{sec:meth_res}, then Section~\ref{sec:concl} concludes this document and presents research directions.

\section{Problem Description}
\label{sec:pb_desc}

\subsection{Context}
The mandate of CBC/Radio-Canada is to inform, enlighten, and entertain all Canadians. When presenting its plan on equity, diversity, and inclusion for 2022-2025, CBC/Radio-Canada committed itself to doing the utmost so that all persons living in Canada feel valued, recognized, and heard by their public broadcaster from sea to sea. 
On its website (radio-canada.ca), Radio-Canada publishes between 450 and 600 articles each day. 
These articles deal with complex topics (health crisis, climate crisis, the economy, the polarization of society, international conflicts, etc.). 
Since a good understanding of current issues is necessary to take part in the democratic debate, Radio-Canada thinks that the use of ATS could help a greater number of citizens take part in this debate. 
Simplifying or summarizing some of its written contents (in an automatic fashion) could enhance the understanding of the articles and make them more attractive for people struggling with literacy, neurodiverse people, and new immigrants (for instance).

In April 2021, CBC/Radio-Canada launched Mauril\footnote{\url{https://mauril.ca/fr/}}, a digital platform for learning French and English through audio-visual information produced by the public broadcaster. The development team is currently trying to broaden the supply of written content through a new reading comprehension task. To this end Radio-Canada will need ATS to produce simpler versions of original articles, in order to take the level of beginners (learning French or English) into account.

\subsection{The Problem}
ATS consists in decreasing the complexity of a text, from the lexical and syntactic points of view, while retaining its meaning and grammaticality. 
This has been shown to improve the readability and ease of understanding of the text. 
Methods for simplifying text fall into two categories: modular systems (which carry out lexical or syntactic simplifying operations iteratively or recursively), and end-to-end systems, which learn to carry out several modifications at the same time through labeled data. In modular systems the transformations are in most cases applied sentence-wise.

\subsection{Desired Solution}
We wish to compare the performance of our modular lexical simplification systems with that of an end-to-end system that operates at the sentence level, without being limited to lexical simplification. The supervised end-to-end approaches that we have identified 
make it possible to perform several types of transformations (both syntactic and lexical), but require a large number of examples. On the other hand, at least two unsupervised approaches~\cite{Martin2020-ot,Surya2019-ri} make it possible to learn how to simplify from free text, without depending on aligned corpora. We also need to identify the appropriate evaluation metrics and evaluation corpora in both French and English. 

\section{Related Work}
\label{sec:rel_work}

Automatic textual simplification consists in reducing the complexity of a text (in terms of lexicon and syntax) while retaining its original meaning, in order to improve its readability and understanding. 
In reality, textual simplification is most often achieved by transformations performed at the sentence level. 
These are rewriting transformations such as replacement (lexical simplification), reorganization (syntactic simplification), and splitting. 
In accordance with the literature, our investigation was centered around sentence-level simplification.

Broadly construed, ATS systems will seek to parse an input sentence and produce an output sentence presumed to be equivalent but linguistically simpler to it. 
As previously mentioned, the inner workings of these systems place them into one of two broad categories: modular and end-to-end systems. 
Modular systems will apply a set of definite, linguistically informed transformations. 
These transformations may pertain to different aspects of language and simplification. 
For example, one transformation may seek to replace words deemed complex with simpler ones, while others will operate on the syntactic structure of sentences. 
Assuming the independence of these aspects as they pertain to simplicity, these transformations will be applied independently by such systems.

In contrast, end-to-end systems will seek to apply such transformations jointly. 
Importantly, said transformations are not necessarily explicitly parameterized by these approaches. 
Instead, end-to-end models will transform the input sentence into a dense vector representation from which the output sentence will be produced: an encoder-decoder model. 
These neural network models will be trained to match the expected output; the simplifying transformations are expected to be learned and effected implicitly through the inner representations of the network.

Thus, end-to-end approaches will usually require training on parallel corpora: datasets matching complex sentences with their simplified versions. 
While it is conceivable to train simple so-called sequence-to-sequence models on this task, their high capacity makes them prone to overfitting on the smaller datasets habitually available in ATS. 
As such, end-to-end approaches based on inference from parallel corpora will attempt to embed structural constraints in the encoding-decoding process in order to rein in model capacity. 
For example, the sentence simplification task can be reframed by having the model issue edit operations to be applied on the input sequence rather than producing the output sequence outright. 
These operations, once interpreted and realized, would result in the simplified sentence. 
For example, the model might be tasked with issuing deletion and preservation operations for each token in the input sentence. 
Carrying out these instructions on each token would result in a simplified sentence. 
The true edit operations constituting the expected output of the model can be produced automatically from the aligned sentence pairs. 
The output space of this modified framework is greatly reduced in dimension, thereby simplifying the task. 
This approach has been exercised in the literature, achieving competitive results~\cite{Dong2019-wu,omelianchuk2021text}. 

Rather than mitigating the impact of limited training examples by reframing the model task, encoder-decoder models can be trained directly to produce simplified text on non-parallel corpora. 
These corpora comprise sets of sentences of each complexity level which are not  individually paired. As such, large amounts of data are more easily gathered. 
Nonetheless, such data do not give a mapping from a complex to a simplified sentence, and the model cannot be trained in a supervised manner. 
\cite{Surya2019-ri} proposed to circumvent this issue by deploying an encoder-decoder model with two decoders, one for each of two complexity levels. 
These decoders are subject to separate training criteria. 
In particular, the decoder producing simple outputs must process both complex sentences and simple sentences in similar manners while doing so differently from the complex decoder. 
In the end, the encoder and simple decoder will constitute a simplification model. 
This approach obtains competitive results on standard datasets. 
However, the authors note that performance can be further improved by additional training on parallel data.

Another approach to addressing limited data is to pose text simplification as a particular case of paraphrasing~\cite{Martin2020-ot}. 
Paired paraphrase examples can be extracted automatically in larger numbers as there is no need for any assessment of their difference in terms of complexity. 
Of course, in order to be used specifically for simplification, the paraphrasing model must be able to be tilted towards producing a simpler paraphrase. 
To this end, the paraphrase model is trained with auxiliary control inputs parameterizing the difference between the input and output sentence with regard to selected aspects~\cite{Martin2020-xv}, e.g. change in total length, change in dependency tree depth. 
Once the model is deemed proficient at paraphrasing, simplifying values of these control inputs are determined empirically from parallel simplification corpora. 
This approach achieves state-of-the-art performance on several datasets as well as good human ratings in English, French and Spanish.

\section{Datasets}
\label{sec:datasets}

ATS corpora vary in several respects, as each attempts to address specific preoccupations of ATS research. 
The modular lexical simplification system developed in this work relies on lexical resources exclusively for evaluation. 
In contrast, the end-to-end sentence simplification approaches studied make use of sentence-level corpora both for training and for evaluation. 
Most resources available of either type are for English; very few resources exist for French.

\subsection{Lexical Simplification}
A multilingual lexical simplification dataset was produced for the TSAR-2022 Shared Task in English, Spanish and Portuguese~\cite{stajner2022lexical}. 
The gold test set in English contains 373 sentences with an identified complex word and multiple simplification suggestions provided by annotators (25 or 26 in some cases). 

To our knowledge, the only evaluation dataset for lexical simplification in French is presented in~\cite{fabre2014presentation}.

\subsection{Sentence Simplification}
The resources of highest interest in the development of sentence-level, end-to-end systems are parallel corpora. 
Broadly, these datasets match equivalent text fragments of different levels of complexity. Nonetheless, they may still vary in several respects. 
A primary distinction is their granularity, for example, whether they focus on sentence- or larger-level simplification. 
Further, datasets may differ in their characterization of the (relative) complexity of the units (e.g. sentences, paragraphs) within each pair thereof. 
For example, in some datasets, observations might consist of two sentences, one deemed complex and the other, simple. 
Alternatively, they may belong to precisely defined categories, borrowed, for example, from Second-Language Acquisition standards. 
This is of particular interest in developing simplification systems capable of producing tailored simplifications Additionally, corpora may have several references per example per target level, offering different, presumably equivalent simplifications of one single sentence. 
This is particularly important in the evaluation of systems, casting a wider net over possible system outputs. 
Finally, some corpora will characterize the differences between units in an observation with respect to several aspects, such as how well the semantic sense is preserved throughout or whether units are grammatically sound.

\subsection{Training Data}
As mentioned, very few resources exist for text simplification in French. 
The only available resources were therefore reserved for evaluation. 
Instead, end-to-end approaches relying on parallel corpora were trained on a machine-translated version of the WikiLarge corpus~\cite{Zhang2017-ri}. 
This dataset is built from sentences extracted from the open collaboration online encyclopedia, Wikipedia, which has both English\footnote{\url{https://en.wikipedia.org/}} and simplified English\footnote{\url{https://simple.wikipedia.org/}} (termed Simple English) versions. 
Sentences were extracted and aligned automatically from articles existing in both versions, for a total of 296k pairs. 
As such, WikiLarge is the largest parallel ATS corpus, to our knowledge. 
Its translation was carried out using the Helsinki machine translation model~\cite{Ostling2017-lu}. 
More details about this process and its considerations are presented in Section~\ref{ssec:datatrans}.

In contrast, the UNTS approach~\cite{Surya2019-ri} does not require a parallel corpus. 
Rather sentences are taken from two separate sets of sentences of the desired levels of complexity, easing the requirements of data collection. 
As such, two sets of data were collected from different sources for the present work. 
Both are news sources, in order to mitigate the impact of domain shift from one set to another. 
Complex sentences were taken from the French-language portion of the MLSUM corpus~\cite{Scialom2020-me}, a multilingual dataset for automatic summarization. 
Simple sentences were collected from the simplified version of the Radio France Internationale (RFI) website\footnote{\url{https://francaisfacile.rfi.fr/}}. 
We term this collection RFI.

The datasets are listed in Table~\ref{tab:datasets}.

\begin{table}[h!]
    \centering
    \renewcommand{\arraystretch}{1.5}
    \begin{tabular}{|l|c|c|l|}
    \hline
    \textbf{Dataset}  & \textbf{Language} & \textbf{Size} & \textbf{Description}\\\hline
    WikiLarge & English & 296K & Automatic alignment of English Wikipedia and Simple English Wikipedia\\\hline
    MLSUM & French & 425K & Multilingual summaries of news\\\hline
    RFI & French & 105K & Transcripts of a news feed in simple French\\\hline
    \end{tabular}
    \vspace{.18cm}
    \caption{Sentence-level resources used in training. WikiLarge was automatically translated into French.}
    \label{tab:datasets}
\end{table}

\subsection{Evaluation data}
Four datasets were considered for the evaluation of end-to-end sentence simplification systems: ALECTOR~\cite{Gala2020-vy}, OneStopEnglish~\cite{Vajjala2018-jk}, TurkCorpus~\cite{Xu2016-yv} and ASSET~\cite{Alva-Manchego2020-hm}. 
The simplifications of all datasets are manual. 
Both ALECTOR and OneStopEnglish comprise document-level simplifications by experts, while TurkCorpus and ASSET comprise sentence-level simplifications by crowdsourced workers. 
The only French-language dataset among those selected is ALECTOR, the rest contain English-language text only. 
Our preliminary results concern only the ALECTOR corpus.

The \textbf{ALECTOR} corpus~\cite{Gala2020-vy} comprises 79 excerpts aimed at children between the ages of 7 and 9, drawn from fiction and scientific texts common to the French curriculum. 
Each excerpt has one associated simplified version aimed at children with dyslexia or poor reading ability. 
Simplifying transformations, carried out by experts, included lexical simplification, deletion and sentence splitting and merging. 
For our purposes, the dataset was adapted by manually aligning sentences. 
Complex excerpts in the original dataset were split at periods and colons, then manually aligned with their simplified counterparts, yielding 1165 pairs. However, because the original simplification was done at the excerpt level, 23\% of pairs have no simplification. 

The \textbf{OneStopEnglish} corpus~\cite{Vajjala2018-jk} contains 2166 groups of sentences taken from OneStopEnglish lesson for English second-language learners. 
The sentences were originally sourced from news articles from The Guardian\footnote{\url{https://www.theguardian.com/}}. 
These articles were adapted for three different reading levels. 
From this text-aligned version, sentence pairs were extracted from each level pair by automatically aligning sentences one-to-one. 
Pairs of sentences deemed too similar or dissimilar were discarded, resulting in different pair counts for each pair of levels.

The \textbf{TurkCorpus}~\cite{Xu2016-yv} comprises 2359 English sentences from Wikipedia for which 8 independent simplifications were produced by different crowdsourced annotators. 
These annotators were instructed to rewrite sentences by substituting challenging words or idioms but without any content deletion or rearrangement, thus limiting the array of simplifying transformations present in the corpus.

To address these limitations, the \textbf{ASSET} corpus~\cite{Alva-Manchego2020-hm} provides 10 simplifications of the same source sentences as the TurkCorpus with a richer set simplifying transformations: paraphrasing, lexical simplification, deletion and sentence splitting and reordering.

\subsection{Dataset Translation}
\label{ssec:datatrans}
\begin{figure}[h!]

\begin{subfigure}{0.5\textwidth}
\includegraphics[width=0.9\linewidth, height=6cm]{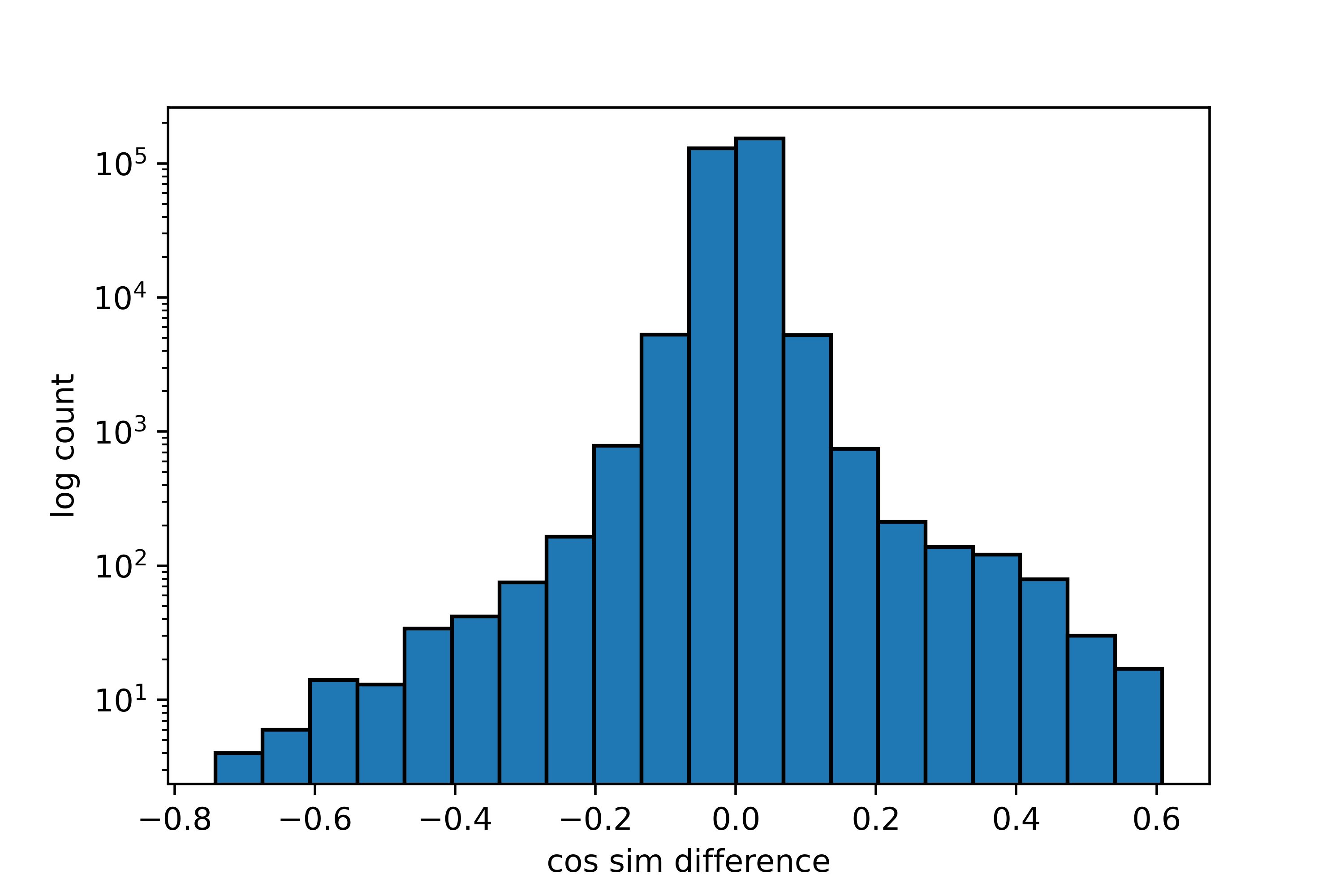}
\end{subfigure}
\begin{subfigure}{0.5\textwidth}
\includegraphics[width=0.9\linewidth, height=6cm]{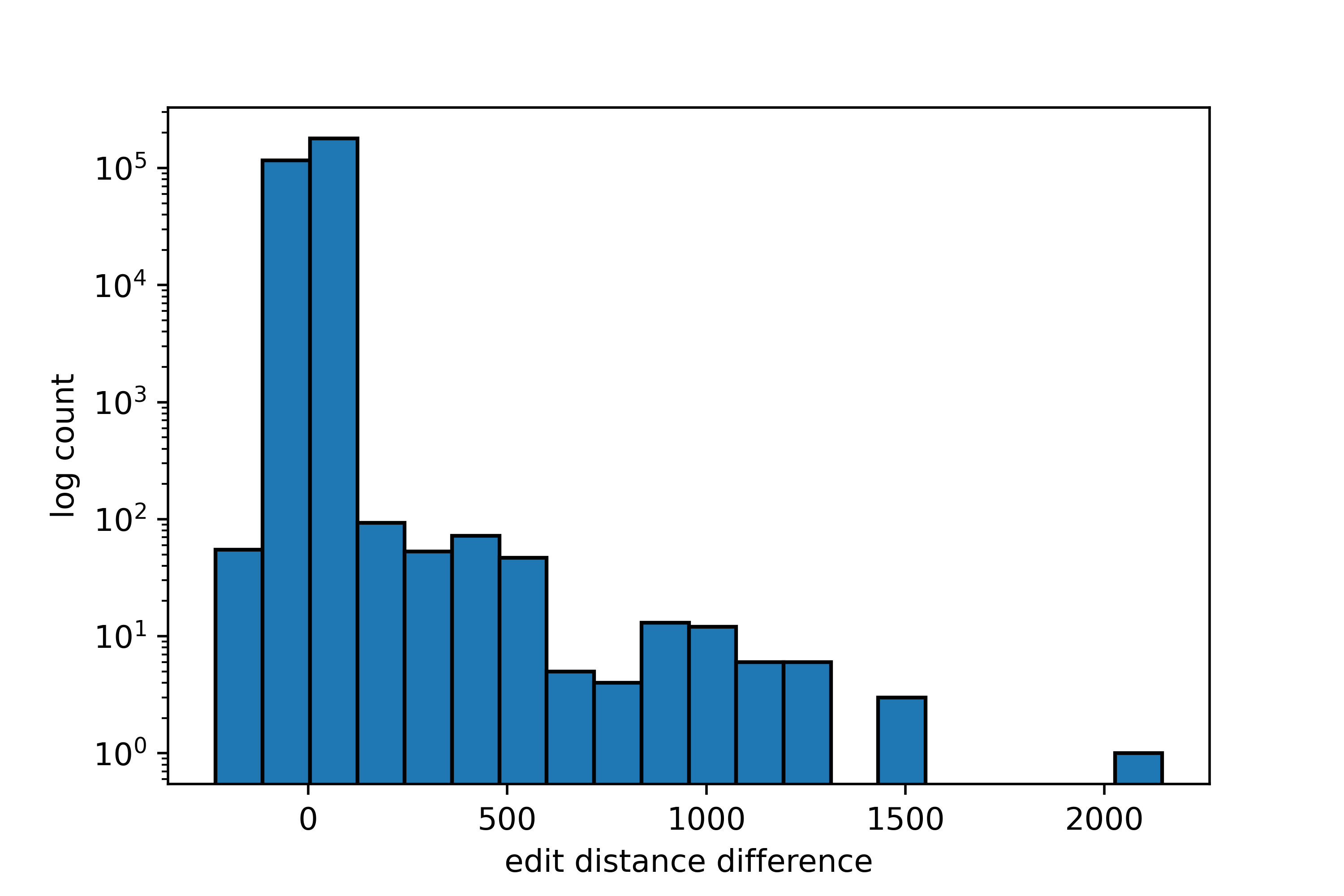}
\end{subfigure}

\caption{Left:~Distribution of difference in cosine similarity between machine-translated (French) and original pairs in the Wikilarge dataset. Right:~Distribution of difference in edit distance.s}
\label{fig:distrib}
\end{figure}

Due to the limited availability of French-language corpora, several datasets were automatically translated from English to French both for training and evaluation purposes. 
This was done using the Helsinki machine translation model~\cite{Ostling2017-lu}. 
Although the quality of machine translation has drastically improved over recent years, it is still important to survey the quality of the translation carried out on the corpora used. 
In particular, it is important to study whether complexity was preserved through translation or, more precisely, whether the discrepancies in complexity within aligned pairs were preserved. 
However, given that measuring linguistic complexity is a difficult problem in itself, certifying this preservation is difficult. 
Little work on these considerations can be found in the literature. 
Namely,~\cite{Rauf2020-no} sought to analyze the effects of translation on the values of surface metrics at different complexity levels. 
Following this idea, we measured changes within sentence pairs of the Wikilarge dataset as per different metrics. 
As a measure of semantic proximity, we computed the cosine similarity between the LASER embeddings~\cite{Artetxe2019-tk} of simple and complex sentences. 
Then the cosine similarity in each original pair was subtracted from that of the translated pair. 
The histograms are shown in Fig.~\ref{fig:distrib}. 
Although most translated pairs have higher similarity than their original counterpart (56\%), the difference in those cases appears to be small (median=0.012). 
However, the syntactic similarity, as approximated by the character-level edit distance, appears to diminish with translation: the median difference among translated pairs that are closer than the original (20\%) is $-4$. 

\section{Evaluation}
\label{sec:eval}

This Section describes the evaluation methods used in the literature for the tasks of lexical and sentence simplification. 
The main components of text simplification are meaning preservation, grammaticality, and simplicity. Usually, human evaluation explicitly evaluates along these while automatic metrics\footnote{In this report, the term “metric” is used in a broader sense than its mathematical definition. Thus, evaluation scores that do not verify the triangular inequality are still referred to as metrics.} can cover multiple aspects. 

\subsection{Human Evaluation}
Because it is difficult to model text simplification with mathematical quantities, human evaluation remains the gold standard. 
This is carried out along the aforementioned features with Likert scales. 
While the evaluation of grammaticality and meaning preservation is fairly straightforward,  there are multiple approaches to the manual evaluation of simplicity. 
Namely, the simplicity of system outputs can be evaluated by themselves~\cite{Stajner2016-ph} or by comparison to the input~\cite{Kriz2020-dk,Sulem2018-zc,Xu2016-yv}.

\subsection{Automatic Evaluation}

\subsubsection{Lexical simplification}
The following evaluation metrics are used in the TSAR-2022 Shared Task for lexical simplification (assuming a correctly identified complex word is provided):
\begin{itemize}
    \item Accuracy@1: the percentage of instances for which the best ranked substitution generated by the system is the same as the most frequently suggested simpler synonym in the gold data.
    \item Mean Average Precision@K: K$\in \{1,3,5,10\}$. MAP@K evaluates the relevance of the predicted substitutes and the position of the relevant candidates compared to the gold annotations.
    \item Potential@K: K$\in \{1,3,5,10\}$. Potential@K evaluates the percentage of instances for which at least one of the substitutions predicted is present in the set of gold annotations.
    \item Accuracy@K@top1: K$\in \{1,2,3\}$. ACC@K@top1 evaluates the ratio of instances where at least one of the K top predicted candidates matches the most frequently suggested substitute in the gold list of annotated candidates.
\end{itemize}

\subsubsection{Sentence simplification}
ATS has been frequently approached as a monolingual translation task~\cite{Dong2019-wu,Zhang2017-ri}. 
That is, complex text is said to be translated to simpler text within the same language. 
This formulation is manifest not only in parallels between approaches but also in their evaluation, with work in the evaluation of ATS systems borrowing from the Machine Translation (MT) literature. 
Some metrics are borrowed without alteration, while other, ATS-specific metrics are more loosely inspired from MT metrics. 
Throughout, the three key dimensions that these metrics seek to capture are:  

\begin{itemize}
    \item grammaticality: whether the output produced by some system is grammatically sound
    \item meaning preservation: whether this output preserves the semantic sense of the input, and 
    \item simplicity: whether the output is indeed simple or simpler than the input. 
\end{itemize}

Five metrics were selected for evaluation in the present work. 

The \textbf{BLEU} metric~\cite{Papineni2002-df}, ubiquitous in MT, is equally common in ATS. 
It has notable weaknesses in this setting, such as its tendency to punish sentence splitting. 
However, it has the advantage of also being usable in a reference-less setting by comparing to the input. 
Naturally, this will reward less intervention. 
In this work, however, BLEU was used only to match against the expected output. 
Another MT metric, \textbf{BERTScore}~\cite{Zhang2020-tr}, relies on pairwise similarities between the token representations of the system output and the references. 
These representations are computed from the BERT language model~\cite{Devlin2019-qz}. 
Basing the comparisons on a large language model rather than symbolic representations allows the metric to account for polysemy and makes it robust to paraphrasing. 
It has shown good correlation with human judgment in ATS tasks~\cite{Scialom2021-el}.

\textbf{SARI}~\cite{Xu2016-yv}, a metric specific to ATS, compares the output to both the reference and the input, in a manner similar to BLEU. 
Although very common in the ATS literature, it is difficult to interpret, and its grasp on grammaticality and meaning preservation remains unclear~\cite{Scialom2021-el}. 
\textbf{SAMSA}~\cite{Sulem2018-zc}, another ATS metric, seeks to measure the simplicity of outputs without the use of references by parsing for the semantic events described therein. 
It thus favors splitting sentences as well as reordering sentences into the dominant sentence structure for the language (Subject-Verb-Object for English). 
Lastly, the \textbf{ISiM}~\cite{Mucida2022-uo}, also reference-less, computes the simplicity of a sentence by relying on the large-scale frequency of words as a proxy for their simplicity, i.e. texts with common words will be deemed simpler regardless of their meaning or lack thereof.

Except for iSiM, all metrics are supported by the EASSE package\footnote{https://github.com/feralvam/easse}~\cite{Alva-Manchego2019-xi}, which was selected for use for this work. 

In future work, we may consider using a measure of similarity (e.g. cosine) between sentence embeddings to capture meaning preservation. 
Further, statistics on the shape of the syntax tree of the outputs of systems (e.g. depth, breadth) could give some insight into the syntactic simplicity of outputs. 
This would improve the above evaluation protocol as its measure of simplicity is quite limited: ISiM, the main metric for this dimension, is oblivious to the adequacy of the simplification and its grammaticality.

It should be noted that automatic evaluation of ATS systems remains an open question~\cite{Alva-Manchego2021-im,Alva-Manchego2020-hm}. 
Most metrics fail to address some aspect of simplification altogether, and may exhibit pitfalls in their evaluation of the aspects they do address. 
In particular, high scores of the above metrics do not imply good simplification quality~\cite{Alva-Manchego2021-im}. 
Furthermore, posing simplicity as solely a textual property without accounting for the competence of the intended audience is an inherently incomplete view of the problem.

\section{ATS Methods and Results}
\label{sec:meth_res}

\subsection{Lexical Simplification}
Our modular system for lexical simplification (for English) requires no training data and allows us to fine-tune each module separately in order to improve the result. 
It consists of three modules operating consecutively: complex word identification (CWI), candidate generation, and candidate ranking. 

\textbf{Complex Word Identification.~~}
To identify the candidate for simplification, we first segment the input sentence into words before passing them through a vocabulary classifier (detailed in the paragraph about \textit{Candidate Ranking}) 
along with the whole sentence as contextual information. The word with the highest complexity score becomes the candidate for simplification. 
In case of a tie, we prioritize verbs and select the word with the highest frequency in order to exclude inappropriate candidates for lexical simplification such as terminology. 

\textbf{Candidate Generation.~~}
To generate substitution candidates, we used the lexical substitution framework \textit{LexSubGen}~\cite{arefyev2022always} and its best performing estimator \textit{XLNet+emb} to generate 20 suggestions given a target complex word. 
We modified the post-processing of the original system to exclude the candidate lemmatization and get inflected suggestions, rather than lemmas. 
We kept the lowercase post-processor followed by target exclusion which uses lemmatization to detect and exclude all forms of the target word. 

\textbf{Candidate Ranking.~~}
\label{sssec:candrank}
We selected and ranked candidates based on a combination of their grammaticality, meaning preservation and simplicity scores through a simple heuristic giving twice as much weight to the simplicity score.
The rank of each substitution $w_n$ of the $N = 20$ generated candidates is determined as a function of its grammaticality $G \in \{0, 1\}$, simplicity $S \in [1, 6]$ and meaning preservation $M \in [1, N]$ scores, as presented in the following equation:

\begin{equation*}
 rank \ \ w_{n\leq N} = G_{w_n} \times (S_{w_n} \times 2 + M_{w_n})   
\end{equation*}

The top 10 ranking candidates (or less) are those included in the submission. 

To evaluate the \textbf{grammaticality} of a sentence given a substitute candidate, we compare the coarse-grained part-of-speech (POS) and morphological features of both complex word and candidate in context. We use spaCy\footnote{\url{https://spacy.io/} | v. 3.1.3 | en-core-web-lg} to tokenize and parse the sentence, making sure not to split hyphenated complex words, since LexSubGen does not support multi-word expressions. We assign a score of 1 to all candidates whose features (person, number, mood, tense, etc.) correspond to those of the target word and 0 otherwise.

To evaluate the effect of a substitution candidate on the \textbf{meaning} of the original sentence we compute the similarity of the two sentences as a sum of the cosine similarities between their token embeddings using BERTScore~\cite{Zhang2020-tr}. 
The higher the similarity between source and target sentences, the higher the chances that the substitution candidate meaning is close to the one of the complex word. 
Candidates are ranked by decreasing F1 score with the best candidate receiving a score of 1 and the last one - a score equal to N.

We measure the \textbf{lexical complexity} of each candidate with a CEFR\footnote{The Common European Framework of Reference for Languages (CEFR) organizes language proficiency in six levels, A1 to C2.} vocabulary classifier trained on data from the English Vocabulary Profile\footnote{\url{https://www.englishprofile.org/american-english}} (EVP)~\cite{capel2012completing}. 
EVP is a rich resource in British and American English which associates single words, phrasal verbs, phrases, and idioms not only with a CEFR level but with part of speech tags, definitions, dictionary examples and examples from learner essays. 
The corpus also contains distinct entries for distinct meanings of polysemous words, each associated with its own difficulty level. 
For each substitution candidate we extract a semantic, contextual, dense vector representation from a pre-trained masked language model\footnote{https://huggingface.co/bert-base-uncased}~\cite{Devlin2019-qz} by first encoding the target word or multi-word expression (MWE) in context (using the dictionary and learner examples) and then aggregating all 12 hidden layers for all WordPieces. This representation of the dataset is then used to train a support vector classifier 5~\cite{platt1999probabilistic}. 
The resulting model is able to assign a difficulty level between 1 (A1) and 6 (C2) to the meaning of any word or MWE as determined by its context.

\textbf{Results and discussion.~~}
We evaluated (partially)\footnote{We did not evaluate the CWI part of the system and leveraged the fact that the TSAR-2022 dataset identifies complex words. 
The results thus presume a perfect complex word identification prior to candidate generation and ranking.} our lexical simplification system (in English) on the TSAR-2022 dataset. 
THe main results are reported in Table~\ref{tab:res}. 
The system outperforms the state-of-the-art \textit{LSBert} baseline on 27 out of the total 51 metrics (including Precision and Recall).

\begin{table}[h!]
    \centering
    \renewcommand{\arraystretch}{1.5}
    \begin{tabular}{|l|c|c|c|c|c|c|}
    \hline
         &  \textbf{ACC@1@top1} & \textbf{ACC@1} &\textbf{MAP@3} & \textbf{Potential@3} & \textbf{Precision@4} & \textbf{Recall@4}\\\hline
    LSBert-baseline & 0.303 & 0.597 & 0.407 & 0.823 & 0.412 & 0.193 \\\hline
    Our system & 0.236 & 0.544 & 0.382 & 0.831 & 0.418 & 0.196\\\hline
    \end{tabular}
        \vspace{.18cm}
    \caption{Results of SOTA and our systems on selected metrics}
    \label{tab:res}
\end{table}

The high Potential@3 suggests that the system would be useful to human editors whenever suggesting 3 (or more) candidates to choose from.
The Accuracy@K@top1 of the system doubles when K = 3 wrt K=1, 
which means that, 46\% of the time, the most commonly suggested substitute is among our top 3 predictions. 
Table~\ref{tab:sentences} contains a number of example simplifications for sentences from the TSAR-2022 corpus.

\begin{table}[h!]
    \centering
    \renewcommand{\arraystretch}{1.5}    
    \begin{tabular}{|l|l|}
    \hline
      \textbf{Sentence with complex word}   & \textbf{Top substitute} \\\hline
       It decomposes to arsenic trioxide, elemental arsenic and iodine when heated in air at 200$^\circ$C.  & changes \\\hline
       Lebanon is sharply split along \textbf{sectarian} lines, with 18 religious sects. & religious \\\hline
       The stretch of DNA transcribed into an RNA molecule is called a transcription unit and \textbf{encodes} at least one gene. & codes \\\hline
        Obama earlier dropped from night skies into Kabul [...], \textbf{cementing} 10 years of U.S. aid for Afghanistan after NATO combat troops leave in 2014. & securing\\\hline       
    \end{tabular}
        \vspace{.18cm}
    \caption{Selected sentences with complex word (in bold) and the top candidate produced by our system}
    \label{tab:sentences}
\end{table}

\subsection{Sentence simplification} 
Having considered the particulars of the problem setting and the needs of CBC/Radio-Canada, namely the lack of French-language resources, several approaches from the literature were considered. 
Selected approaches were UNTS~\cite{Surya2019-ri}, ACCESS~\cite{Martin2020-xv}, MUSS~\cite{Martin2020-ot}. 
Our preliminary results concern only the Alector corpus and the BLEU and SARI metrics. 
UNTS was trained on the MLSUM and RFI datasets; ACCESS was trained on the translated Wikilarge corpus. 
MUSS was used as trained by its original authors~\cite{Martin2020-ot}. 
They are presented in Table~\ref{tab:modelres}.

\begin{table}[h!]
    \centering
    \renewcommand{\arraystretch}{1.5}
    \begin{tabular}{|l|c|c|}
    \hline
     \textbf{Model}    &  \textbf{BLEU} & \textbf{SARI}\\\hline
      UNTS   &  37.6 & 35.5 \\\hline
      ACCESS & 46.7 & 34.9  \\\hline
      MUSS   & 38.9 & 38.1 \\\hline      
    \end{tabular}
    \vspace{.18cm}
    \caption{Results of selected models on the sentence-aligned French-language Alector corpus. }
    \label{tab:modelres}
\end{table}

The paraphrase-based approach, MUSS achieves the highest SARI score, while ACCESS achieves the highest BLEU but at the cost of a lower SARI. 
This is consistent with the reported state-of-the-art performance of MUSS on other datasets~\cite{Martin2020-ot}. 
However, it should be noted that these results are fragile in that Alector only has a single reference output whereas both metrics are more reliable with several reference outputs~\cite{Alva-Manchego2021-im,Xu2016-yv}. 
UNTS was trained on the MLSUM and RFI datasets; ACCESS was trained on the translated Wikilarge corpus. MUSS was used as trained by its original authors. 

\section{Conclusion}
\label{sec:concl}

Evaluation for ATS tasks is not standard and the current metrics have known limitations~\cite{Alva-Manchego2021-im}. 
In this work, a protocol to evaluate our ATS systems was identified. 
It comprises multiple metrics to capture the 3 dimensions of text simplification: grammaticality, meaning preservation, and simplicity. 
While imperfect, this metric ensemble allows for deeper analyses of our systems.

This work explored two avenues for text simplification: lexical simplification and end-to-end systems for sentence simplification. 
Our lexical simplification solution obtained similar performance as those of the shared task baseline~\cite{Stajner2016-ph} and showed promising results when considering the top 3 to 4 words. 
This suggests that it could be a good system to support expert annotators. 
Additionally, this report presents partial results for the ALECTOR dataset~\cite{Gala2020-vy} with several end-to-end systems. 
While fragile due to the lack of multiple references, results of ACCESS~\cite{Martin2020-xv} and MUSS~\cite{Martin2020-ot}, an ACCESS-based approach, suggest that further adjustments of this controllable model could further improve simplification quality.

To complete this work, a deeper evaluation of the presented approaches along the 3 ATS-dimensions – meaning preservation, grammaticality, and simplicity – is needed to appropriately assess the quality of the discussed systems. 
Moreover, this evaluation could be complemented by the use of translated multiple-reference corpora such as ASSET~\cite{Alva-Manchego2020-hm}.

\bibliographystyle{unsrtnat}
\bibliography{ARPI2022_ATS_RadioCanada_UQAM}

\end{document}